\documentclass[sigconf]{acmart}
\AtBeginDocument{%
  }

\copyrightyear{2027}
\acmYear{2027}
\setcopyright{cc}
\setcctype{by}
\acmConference[WSDM '27]{Proceedings of the Twentieth ACM International Conference on Web Search and Data Mining}{February 15--19, 2027}{Hong Kong, Hong Kong}
\acmBooktitle{Proceedings of the Twentieth ACM International Conference on Web Search and Data Mining (WSDM '27), February 15--19, 2027, Hong Kong, Hong Kong}
\acmDOI{10.1145/3862322.3872125}
\acmISBN{979-8-4007-3117-4/2027/02}
\newcommand{\grader}{CHiLLER}

\begin{document}

\title{\grader: Calibrated Human-in-the-Loop Short-Answer\\ Grading with Grounded Feedback}

\author{Pranav Raikote}
\correspondingauthor
\email{pranav.raikote@dsv.su.se}
\orcid{0009-0008-7464-5828}
\affiliation{%
  \institution{Stockholm University}
  \city{Stockholm}
  \country{Sweden}
}

\author{Korbinian Randl}
\email{korbinian.randl@dsv.su.se}
\orcid{0000-0002-7938-2747}
\affiliation{%
  \institution{Stockholm University}
  \city{Stockholm}
  \country{Sweden}
}

\author{Ioanna Miliou}
\email{ioanna.miliou@dsv.su.se}
\orcid{0000-0002-1357-1967}
\affiliation{%
 \institution{Stockholm University}
 \city{Stockholm}
 \country{Sweden}}

\author{Panagiotis Papapetrou}
\email{panagiotis@dsv.su.se}
\orcid{0000-0002-4632-4815}
\affiliation{%
  \institution{Stockholm University}
  \city{Stockholm}
  \country{Sweden}}

\renewcommand{\shortauthors}{Raikote et al.}

\begin{abstract}
Large Language Models (LLMs) offer promising capabilities for automated grading, but their adoption in educational assessment raises concerns about grading reliability, transparency, and student data privacy. We present \grader{}, an on-premise, human-in-the-loop grading system built on open-weight LLMs. The system combines calibrated confidence-based routing, which selectively defers uncertain predictions to teachers, with gradient-based token attribution and retrieval-grounded feedback from course materials. Teachers can inspect predictions, adjust grades, and provide corrections for subsequent model adaptation. Evaluated on 180 held-out student answers, the system routes only 16.1\% of answers for human review while capturing 75\% of large grading errors. By integrating selective grading, explainability, and source-grounded feedback in a locally deployable architecture, \grader{} offers a practical approach to AI-assisted assessment that preserves teacher oversight and avoids transmitting student data to external providers.
\end{abstract}




\maketitle

\section{Introduction}

As university courses grow in size, teachers face increasing pressure to assess students both fairly and efficiently. Short-answer questions are well-suited to assessing conceptual understanding and reasoning, but grading them by hand is time-consuming. Manual grading is also prone to inconsistency, as fatigue, the order in which answers are read, and variations in the application of grading criteria over a long grading session can lead to different grades for answers of similar quality.

LLM-based grading systems can considerably reduce this burden, as their agreement with human graders on this task is approaching inter-human agreement levels ~\cite{fredericeneye2025survey}. However, high grading agreement does not necessarily guarantee reliable grading of individual answers.
LLM-based graders can assign incorrect grades with high confidence making such errors difficult to spot. Moreover, their stochastic nature can lead to different grades for the same answer across repeated runs. Such inconsistencies can affect a student's grade on individual questions and, in turn, on the whole examination, which calls for keeping the teacher in the grading loop.

\begin{table}[t]
\caption{Related LLM-based grading systems. Routing: grades are accepted or routed by per-answer confidence; Open: open-weight models; Retrieval: grading or feedback uses retrieved course material. Cites: feedback cites its sources. \checkmark{} yes, $\sim$ partial, $\times$ no or not reported. $^\dagger$Routes items disputed by students.}
\label{tab:compare}
\small\setlength{\tabcolsep}{3.5pt}
\begin{tabular}{lcccc}
\toprule
& Routing & Open & Retrieval & Cites \\
\midrule
\multicolumn{5}{l}{\textit{Used with students}} \\
Pensieve~\cite{yang2025pensieve}                   & \checkmark & $\times$   & $\times$   & $\times$ \\
Avalon~\cite{armfield2025avalon}                   & $\sim^\dagger$ & $\times$ & $\times$ & $\times$ \\
Vanhoyweghen et al.~\cite{vanhoyweghen2026hitl}    & $\sim$     & $\times$   & $\times$   & $\times$ \\
Vangelova et al.~\cite{vangelova2026moodle}        & $\times$   & $\times$   & \checkmark & $\times$ \\
\midrule
\multicolumn{5}{l}{\textit{Offline: confidence for LLM grading}} \\
SURE~\cite{korthals2026sure}                       & \checkmark & $\sim$     & $\times$   & $\times$ \\
Deng et al.~\cite{deng2026rubric}                  & \checkmark & \checkmark & $\times$   & $\times$ \\
Ferrer et al.~\cite{ferrer2026trust}               & \checkmark & \checkmark & $\times$   & $\times$ \\
\midrule
\multicolumn{5}{l}{\textit{Offline: retrieval for grading}} \\
Chu et al.~\cite{chu2025rag}                       & $\times$   & $\times$   & \checkmark & $\times$ \\
Xu \& Corter~\cite{xu2026graphrag}                 & $\times$   & $\times$   & \checkmark & \checkmark \\
\midrule
\textbf{\grader{}}                             & \checkmark & \checkmark & \checkmark & \checkmark \\
\bottomrule
\end{tabular}
\end{table}

Existing systems do so by having the teacher review every grade produced by the system~\cite{lin2026verigrader}, review items that students dispute~\cite{armfield2025avalon},
or review predictions about which the system is uncertain~\cite{korthals2026sure, yang2025pensieve}. Most of them rely on commercial APIs for at least one component of the grading systems. This creates two additional problems. Firstly, sending student answers to an external provider is difficult to align with the Education, Ethics, and Privacy policies~\cite{euai2026guidelines} and the requirements placed on systems evaluating learning outcomes. Furthermore, closed models offer no access to weights or gradients, and are increasingly restricting fine-tuning~\cite{openai2026deprecations}; therefore, the LLM-based grader cannot be inspected through attribution nor adapted to a course, and the teacher is left with the system's own explanation of its grade, which is often unfaithful~\cite{madsen2024selfexplanations}.

To address these challenges, we propose \grader{}, a Human-in-the-Loop (HiL) grading system that runs entirely on open-weight LLM models on local university hardware. To position \grader{} within the broader landscape of LLM-based grading, Table~\ref{tab:compare} compares representative systems against  across four dimensions: confidence based routing, open-weight models, retrieval of course materials, and source-grounded feedback. Systems deployed in educational settings predominantly rely on commercial models and rarely incorporate retrieval of course materials. Pensieve~\cite{yang2025pensieve} is the closest to our approach in its use of confidence-based routing to teachers. Open-weight models have primarily been investigated in offline studies of grading confidence~\cite{korthals2026sure, deng2026rubric, ferrer2026trust}, while retrieval-based grading has likewise largely been evaluated offline~\cite{chu2025rag, xu2026graphrag}. 

To the best of our knowledge, no prior system combines all four capabilities in a unified, locally deployed grading framework.
Building on the calibrated routing method of our earlier work~\cite{raikote2026chillgrader}, the contributions of \grader{} include: \textbf{(i)}~a per-exam workflow in which the teacher grades a small seed batch of the current examination, which is used to adapt the LLM-based grader and calibrate its confidence, and then sets how much of the remaining grading is delegated; \textbf{(ii)}~open-weight models that can be fine-tuned on local hardware in minutes and expose gradients for token attribution, making the grader inspectable; and \textbf{(iii)}~a retrieval-augmented feedback module that generates feedback for all student answers, citing sources from the lecture materials. The source code is available in GitHub\footnote{\url{https://github.com/pranavraikote/chiller}} and the system is publicly available\footnote{\url{https://chillgrader-preview.dsv.su.se/}}.




\section{System Architecture}
\grader{} follows a HiL grading workflow that combines four components: \textbf{(1)}~calibrated confidence-based grading, \textbf{(2)}~selective routing of uncertain predictions to teachers,  \textbf{(3)}~attribution to support grade review, and \textbf{(4)}~retrieval-grounded feedback. Teacher corrections are recorded for subsequent model adaptation.


\paragraph{Calibrated Grading}
\label{sec:sys_arch}

The grading component uses Qwen2.5-7B-Instruct \cite{qwen2025qwen25}, fine-tuned on graded answers from the 2021--2023 editions of \textit{DAMI}, a master's level data mining course, and further adapted to a teacher-graded seed batch of its Spring 2025 examination, the \emph{target} examination using low-rank adaptation~(LoRA)~\cite{hu2022lora}. Following~\cite{raikote2026chillgrader}, rather than directly generating a grade, the grader scores the structured response corresponding to each admissible grade $y\in\{0,\dots, G\}$ (i.e.,``\texttt{\{"grade": $y$, "max\_grade": $G$\}}''). We denote its log-likelihood $\ell(y\mid x)$. Treating these scores as logits, temperature scaling~\cite{guo2017calibration} yields the distribution,
\begin{equation}
    p_T(y\mid x)=\frac{\exp(\ell(y\mid x)/T)}{\sum_{y'}\exp(\ell(y'\mid x)/T)}
\end{equation}
from which we take the predicted grade $\hat{y}=\arg\max_y p_T(y\mid x)$ and its confidence $c(x)=\max_y p_T(y\mid x)$. 
We calibrate the confidence scores using temperature scaling. The temperature $T$ is fit on a teacher-graded validation split by minimizing expected calibration error (ECE), calibrating confidence $c(x)$ without changing $\hat{y}$.

\paragraph{Confidence-Based Routing}
The calibrated confidence controls whether a prediction is accepted automatically or deferred to the teacher. Given a threshold $\tau$, predictions with $c(x)\geq\tau$ are \texttt{accepted}; otherwise the grader abstains and the answer is \texttt{routed} for manual grading. We use $\tau=0.50$ as the operating point, fixed before test evaluation and adjustable in the interface. Since candidate-grade scoring yields both $\hat{y}$ and $c(x)$, routing adds no extra inference cost, unlike confidence estimation via repeated sampling~\cite{korthals2026sure, vanhoyweghen2026hitl} or by generating self-explanations~\cite{cong2026confidence}.


\begin{figure*}
    \centering
    \includegraphics[width=18cm, height=8cm, keepaspectratio]{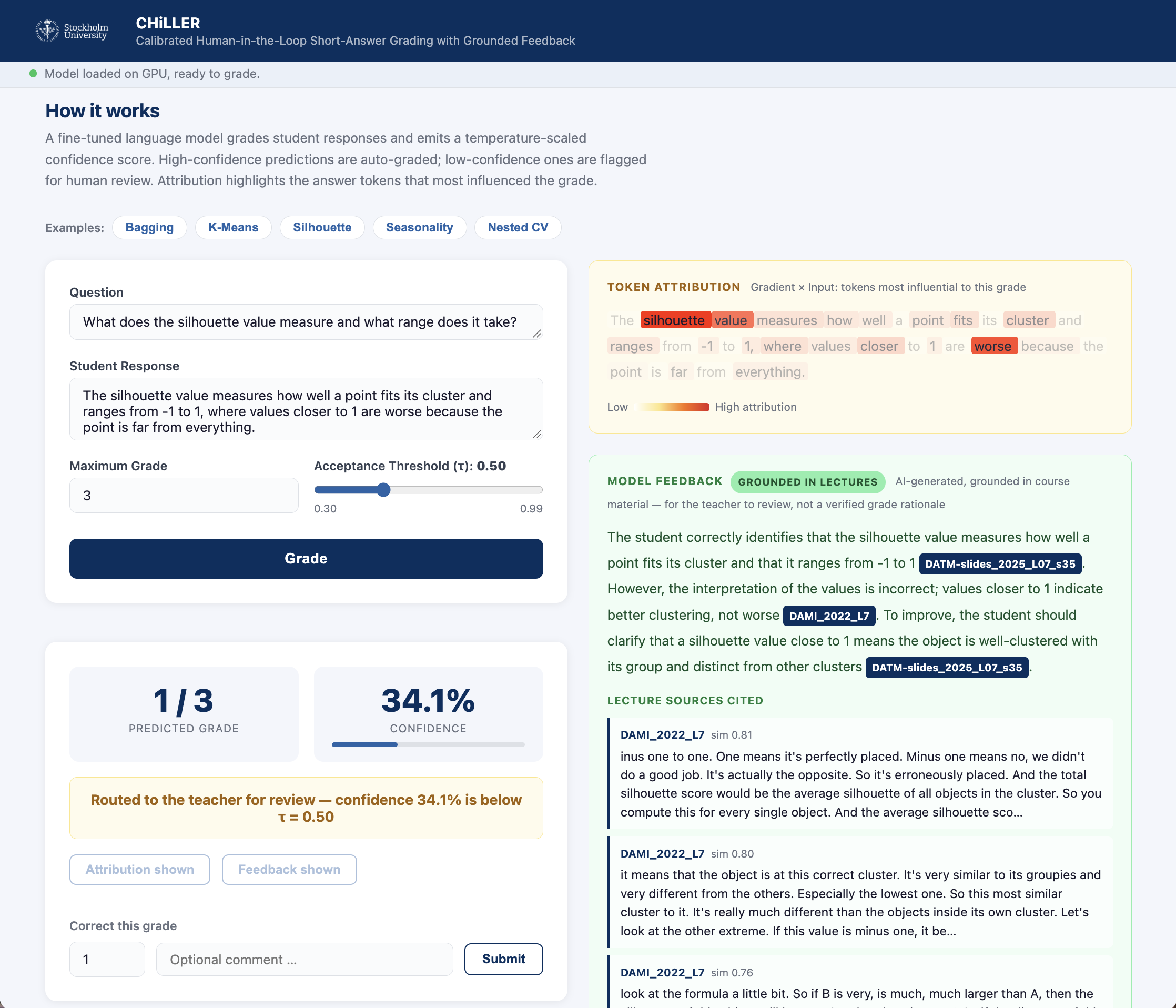}
    \caption{A routed answer. The grader predicts 1 out of 3 marks with confidence 0.34, which is below $\tau=0.50$. The answer is routed to the teacher for review; the teacher sees which words influenced the grade and the feedback generated, and can correct or confirm the grade with a comment.}
    \Description{Screenshot of a two-column web interface. The left column holds the question, the student answer, a maximum grade of 3 and a predicted grade of 1/3, a threshold slider set to 0.50, a confidence of 44.2\%. A yellow notice that shows the answer is routed to the teacher, a box for entering a corrected grade, and a comment. The right column shows the student answer with words shaded from light to dark red by importance, and then a green panel with generated feedback containing inline slide labels, and below it the text of a cited slide with its similarity score.}
    \label{fig:placeholder}
\end{figure*}

\paragraph{Token Attribution} To show which parts of an answer most influence its grade, the system computes the Gradient$\times$Input attributions~\cite{shrikumar2017learning} with respect to the log-likelihood of the predicted grade response, $\ell(\hat{y}\mid x)=\sum_t \log p(\hat{y}_t\mid x,\hat{y}_{<t})$. For each token $i$ with input embedding $e_i$, the element-wise product of the embedding and its gradient is aggregated with the L2 norm over the embedding dimension,
\begin{equation}
s_i=\bigl\lVert e_i\odot\nabla_{e_i}\,\ell(\hat{y}\mid x)\bigr\rVert_2
\end{equation}

which yields a non-negative importance score from a single backward pass. Unlike commercial API-based models, this approach requires access to the model weights. Furthermore, the sub-word scores are averaged into word scores, while stop words and punctuation are removed for the user interface.

\paragraph{Retrieval-grounded Feedback} The retrieval index contains chunks from the lecture material of 2021--2023 and 2025 editions of the course, with one chunk per slide and transcripts split into segments of about 700 characters. Chunks are embedded with the Sentence-BERT~\cite{reimers2019sbert}, normalized to unit length, and stored in a FAISS index~\cite{johnson2021faiss}. The question and answer together form a query, and the $k=5$ most similar chunks are retrieved~\cite{lewis2020rag}, with minimum slots reserved for both slides and transcript. A separate feedback model, Qwen2.5-14B-Instruct~\cite{qwen2025qwen25}, is instructed to generate the feedback by first acknowledging what the answer covers, then names the concepts missing for full marks. The feedback model is also instructed to strictly cite the slide and transcript chunks. 

If the highest similarity falls below $\theta=0.35$, feedback is generated without retrieved context and marked as \texttt{uncited}. We set $\theta$ by inspecting the retrieval scores on probe answers: on-topic answers reach a top-1 similarity of about 0.6--0.8, and the lowest top-1 on the studied examination was 0.40. Feedback can be requested for any answer and is reviewed by the teacher before release to students.

\paragraph{Corrections} When a teacher corrects or updates a grade, the predicted grade, the corrected grade, and an optional comment are written to a correction pool, which serves as training data for subsequent adapters.

\section{Demonstration and Results}
\label{sec:past_exam}

As illustrated in Figure~\ref{fig:placeholder}, we demonstrate the system through an interactive interface that allows teachers to submit questions and student answers, obtain the predicted grade and confidence scores, and optionally adjust the routing threshold. 
Depending on the confidence, an answer is \texttt{accepted} or \texttt{routed} to the teacher for review. Teachers can also inspect token attributions, access feedback grounded in lecture materials, and correct grades or feedback before releasing them. The demonstration uses questions from the target examination and illustrative answers prepared by the authors.

\paragraph{Setup} 
To evaluate this workflow, we replay it on the Spring 2025 version of the course; the first application of this system used 976 answers graded across 45 questions, out of which 657 answers acted as the seed batch for fine-tuning, 139 for validation, and 180 for testing, with 43 of the 45 questions appearing in the test split. We report agreement within 1 mark (when the predicted and true grade differ by at most one point), and define deviations of two or more marks as large errors.

\paragraph{Routing}

\begin{table}[t]
\caption{Routing on 180 held-out answers. Large errors deviate from the teacher's grade by two or more marks (12 in total). $\tau=0$ accepts every grade; the deployed threshold is $\tau=0.50$.}
\label{tab:routing}
\small
\begin{tabular}{lcccccc}
\toprule
 & \multicolumn{3}{c}{Accepted} & \multicolumn{3}{c}{Routed to teacher} \\
\cmidrule(lr){2-4}\cmidrule(lr){5-7}
$\tau$ & $n$ & Exact & Within-1 & $n$ & Exact & Large err. \\
\midrule
0             & 180 & 74.4 & 93.3 & 0  & --   & 0/12 \\
0.45          & 161 & 78.9 & 97.5 & 19 & 36.8 & 8/12 \\
\textbf{0.50} & 151 & 82.1 & 98.0 & 29 & 34.5 & 9/12 \\
0.60          & 127 & 86.6 & 99.2 & 53 & 45.3 & 11/12 \\
\bottomrule
\end{tabular}
\end{table}

At the deployed threshold, 29 of the 180 answers are routed to the teacher, and they contain 9 of the 12 large errors as shown in Table ~\ref{tab:routing}. Exact agreement is 34.5\% on the routed answers and 82.1\% on the accepted ones, so the teacher's review falls specifically on the answers the grader struggles to be confident in its grade predictions. Lowering $\tau$ to 0.45 routes 11\% of the answers and 8 of the large errors, while raising it to 0.60 routes 29\%. This lets the teacher trade review effort against agreement.

\paragraph{Efficiency and Cost}

Table~\ref{tab:cost} summarizes the resources required for full examination; grading and feedback run on separate GPUs. Adapting the grader took about 16 minutes on a single GPU, and grading plus feedback for the routed answers required 1.02 GPU-hours. Although the monetary difference to a single API call is modest at the scale of one examination, confidence estimation either from repeated sampling or self-explanations multiplies the cost and compute, whereas our system obtains it from the same scoring step as the grade. Moreover, neither adaptation nor gradient-based attribution is available without access to model weights. 

\begin{table}[t]
\caption{Efficiency and cost for the examination of 976 answers. On-premise figures were measured on RTX A5500 GPUs; API costs use list prices as of October 2026 and our measured prompt lengths, excluding reasoning tokens.}
\label{tab:cost}
\small
\begin{tabular}{lrrr}
\toprule
 & \multicolumn{1}{c}{Our System} & \multicolumn{1}{c}{GPT-5.4 mini} & \multicolumn{1}{c}{GPT-5.4} \\
\midrule
Grading, single pass        & 0.06 GPU-h  & \$0.17 & \$0.58 \\
Feedback (routed)           & 0.96 GPU-h  & \$0.27 & \$0.91 \\
Adaptation to the course    & 0.27 GPU-h  & N/A$^{*}$ & N/A$^{*}$ \\
Token attribution           & 124 ms/answer & N/A & N/A \\
\midrule
Latency, grade / feedback   & 0.23 s / 22 s & & \\
Memory, grader / feedback   & 16.5 / 17.6 GB & & \\
\bottomrule
\end{tabular}

\smallskip
\footnotesize $^{*}$Self-serve fine-tuning is no longer offered to organizations~\cite{openai2026deprecations}.
\end{table}


\paragraph{Discussion}
Developing \grader{} provided us with three practical insights.
First, separating grading from feedback generation proved
valuable. While the fine-tuned grader was effective at
predicting grades, its generated feedback was often
unreliable or hallucinated. Using a dedicated model for
feedback generation allowed us to specialize the two
components for their respective tasks and ground feedback
in retrieved course materials.

Second, confidence calibration was strongly course-dependent.
A temperature calibrated on earlier editions of the course produced
overconfident predictions on the target examination,
highlighting the importance of adapting and calibrating
the grader using representative teacher-graded answers.

Third, our prompt-injection experiments revealed that
high confidence does not necessarily imply a trustworthy
prediction. An insufficiently adapted grader
assigned full marks with maximum confidence to an answer
explicitly requesting them. This reinforced the importance
of combining selective automation with teacher oversight,
rather than treating confidence-based routing as a
guarantee of grading correctness.

\section{Conclusion}
We presented \grader{}, an on-premise, human-in-the-loop grading system that combines calibrated grading, confidence-based routing, gradient-based attribution, and retrieval-grounded feedback using open-weight LLMs. By selectively routing uncertain predictions to teachers, the system aims to reduce manual grading effort while preserving human oversight and student data privacy. Our evaluation on 180 held-out answers shows that routing only 16.1\% of answers to teachers captures 75\% of large grading errors, demonstrating the potential of selective grading to focus human attention where it is most needed. Future work will investigate iterative adaptation from teacher corrections, robustness across courses and student cohorts, and the system's effectiveness in real-world examinations.

\begin{acks}

This research is funded by the European Health and Digital Executive Agency (HADEA) for the project AI and Health (Grant Agreement 101083880). Views and opinions expressed are, however, those of the author(s) only and do not necessarily reflect those of the European Union or the European Research Executive Agency.
\end{acks}



\bibliographystyle{ACM-Reference-Format}
\bibliography{sample-base}









\end{document}